# Sample-specific repetitive learning for photo aesthetic assessment and highlight region extraction

Ying Dai

Iwate prefectural university

dai@iwate-pu.ac.jp

**Abstract**

Aesthetic assessment is subjective, and the distribution of the aesthetic levels is imbalanced. In order to realize the auto-assessment of photo aesthetics, we focus on retraining the CNN-based aesthetic assessment model by dropping out the unavailable samples in the middle levels from the training data set repetitively to overcome the effect of imbalanced aesthetic data on classification. Further, the method of extracting aesthetics highlight region of the photo image by using the two repetitively trained models is presented. Therefore, the correlation of the extracted region with the aesthetic levels is analyzed to illustrate what aesthetics features influence the aesthetic quality of the photo. Moreover, the testing data set is from the different data source called 500px. Experimental results show that the proposed method is effective.

## 1. Introduction

In response to the growth of digital camera, more and more pictures are taken to upload the social media. Many people hope to improve aesthetic level of themselves by taking beautiful photographs. So, auto-assessment of photo aesthetics is challenging. Researches have been investigating methods for providing automated aesthetical evaluation and classification of photographs. Aesthetic assessment is subjective. One of the main difficulties in addressing this challenge is in developing formal models of human aesthetic preference [1]. In this paper, authors stated that such models would allow computer systems to predict the aesthetic taste of a human being or adapt to the aesthetic tendencies of a human group. For making aesthetics automatic evaluation and choices, the best way to proceed is to create datasets for training the model in collaboration psychology aesthetics (PA) researchers, because computational aesthetics (CA) research typically reposts results using a success rate, while psychologists are more likely to use correlation. Closer collaboration between CA and PA can give rise to results that advance both disciplines. In [2], recent computer vision techniques used in the assessment of image aesthetic quality were reviewed. The authors discussed the possibility of manipulating the aesthetics of images through computational approaches. The research reviewed in the paper generally aims at assessing the aesthetic quality of photos with aesthetic scores or distinguishing high-quality photos form low-quality photos, by training the photo aesthetic models based on the deep learning techniques. However, such models can't interpret which salient image composition features and highlight regions are correlated with the photo aesthetics. Moreover, who labeled the aesthetic scores of training data set for deep learning, professional photographer or amateur, is unclear. In [3], a set of features derived from both low- and high-level analysis of photo layout were exploited to perform the aesthetic quality evaluation by a Support Vector Machine (SVM) classifier. In [4], authors designed a set of compact rule-based features based on photographic rules and aesthetic attributes, and used Deep Convolutional Neural Network (DCNN) descriptor to implicitly describe the photo quality. These approaches focused on extracting the handcrafted image features. However, the effectiveness is limited that extracting the features is based on the researchers' understanding on the aesthetic rules. In [5], the images were divided into three categories: "scene", "object" and "texture". Each category has an associated convolutional neural network (CNN) which learns the aesthetic features for the category classification. In [6], a scene convolutional layer was designed to learn specific aesthetic features for various scenes by deep learning model. In [7], a novel photograph aesthetic classifier with a deep and wide CNN for fine-granularity aesthetical quality prediction was introduced. However, the correlation of the extracted features with the photo aesthetic assessment was not interpreted in the view of PA in such research. In [8], the percentage distributions for orientation, curvature, color and global symmetry were extracted and fed to a deep neural network under the form of only 114 inputs. Differences in extracted features between aesthetically good and poor images were analyzed and some human aesthetic preferences in static two-dimensional scenes were observed. However, the issue whether the handcrafted features are generic for the photo aesthetic assessment is not involved. Moreover, all of the above approaches were not involved in the issue that the aesthetic rating is ambiguous and is different from person to person, which caused a highly imbalanced distribution of aesthetic ratings. Toward to tackling these issues, authors in [9] showed how to learn deep features for imbalanced data classification. Using the learned features, the classification was simply achieved by a fast cluster-wise kNN search followed by a local large margin decision. In [10], authors proposed an end-to-end CNN model which simultaneously implements aesthetic classification and understanding. A sample-specific classification method that re-weights samples' importance is implemented, and what is learned in the deep model was investigated. Ambiguous samples are given lower weights while clear samples are weighted high. However, the method to give the weight of every sample was not explicit, and the improvement for the imbalanced data classification was not salient from the experiment results. Further, the correlation of the learned deep features with the aesthetic assessment was not analyzed although deep activation map was visualized.

Motivated by the above research, we collected about 3100 photos scored aesthetically by a professional photographer who could be considered as a PA researcher. These photos were taken by the students of the photographer's class. The scores are in the range of [2, 9]. The photos with score 2 or less are aesthetically poor; those with score 3 to 4 are fair; those with score 5 and 6 are good; those with 7 or more are excellent. The data set indeed exhibited a highly non-uniform distribution over score as illustrated in Fig.1. The majority images concentrate on the values of 3 to 5 (more than 80%). The model could be overwhelmed by those general samples if the parameters are learned by treating all samples equally, and the more salient samples couldn't decide how the model is trained.

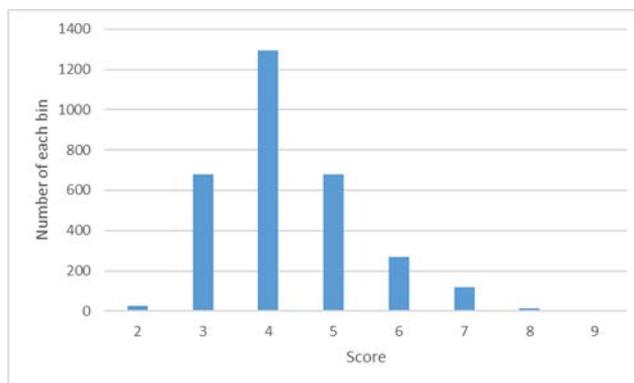

Fig. 1 Score distribution of photo data set

Accordingly, in order to solve the data imbalance issue in aesthetic assessment, in this paper, we focus on training the convolutional neural network (CNN)-based aesthetic assessment model by dropping out the unavailable samples scored in the range of [3, 5] from the training data set repetitively to overcome the effect of imbalanced aesthetic data on classification. Further, the aesthetic highlight region of photo images are extracted by subtracting two specific feature maps of first convolutional layers of two repetitively trained models, to analyze the correlations of the highlight regions with the aesthetic assessments, and explain what aesthetic information influence the aesthetic quality of the photos. Furthermore, the source of testing data set is different from one of the training data set. It was collected from the recommended photos of 500px which is an online photography network. The experimental results show that the proposed method is effective.

## 2. Related work

The photos' aesthetic level assessment exhibit highly-skewed score distribution as shown in Fig. 1. As described in [9], for such class-imbalanced data, the minority class often contains very few instances with high degree of visual variability. The scarcity and high variability make the genuine neighborhood of these instances easy to be invaded by other imposter nearest neighbors. In [11], a comprehensive literature survey to tackle the class data imbalance problem was reviewed. Generally, there are two groups of solutions: data re-sampling and cost-sensitive learning. The former group focuses on learning equally good classifiers by random under-sampling and over-sampling techniques. The latter group operates at the algorithmic level by adjusting misclassification. A well-known issue with over-sampling is its tendency to overfitting. Therefore, under-sampling is often preferred, although potentially valuable information may be removed. Cost-sensitive alternatives avoid these problems by directly imposing heavier penalty on misclassifying the minority class. In [9], a data structure-aware deep learning approach with build-in margins for imbalanced classification was proposed. However, these literature methods mainly aim at the classification of the classes which are defined explicitly. For the classification of aesthetic assessment scores which are of ambiguousness, the state-of-art methods seem not to be available.

## 3. Training photo aesthetic assessment model

In this paper, we propose a CNN-based learning method for photo Aesthetic assessment model by repetitively dropping out the low likelihood samples of majority score classes, so as to ameliorate the invasion to the minority score classes, and the loss of valuable features of majority instances. The idea behind is the assumption that the instance with low likelihood to the majority score classes is what is ambiguously assessed. These samples are easy to overwhelm the Aesthetic assessment model. The system diagram is shown in Fig. 2.

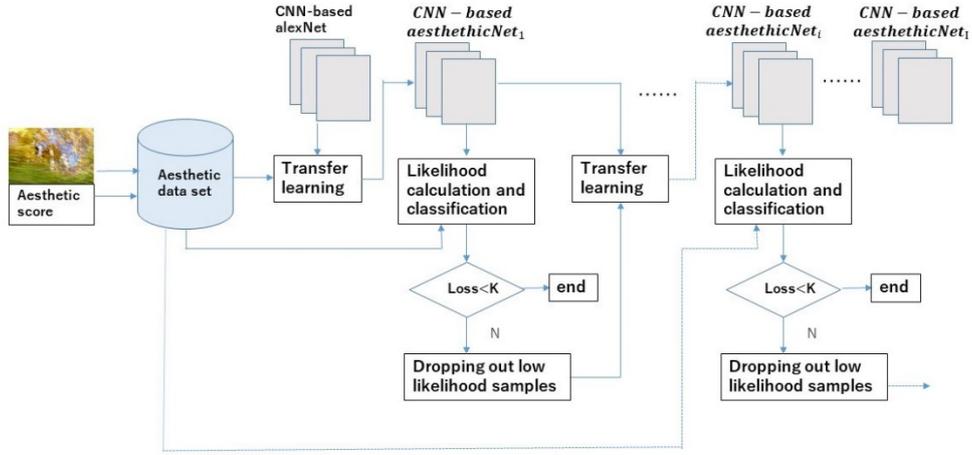

Fig. 2 Diagram of training photo Aesthetic assessment model

The training data set is the score-labeled photo data set. The aesthetic scores in the range of [1, N] are given by a pro-photographer. The scores' distribution of the collected photos is as shown in Fig. 1. The samples' scores are almost concentrated in the midlevel classes. Score $n$ is handled as one class which is denoted $s_n$, while the number of classes is $N$.

The photo aesthetic assessment model is tackled to be a CNN-based classification network of ambiguous classes. Training begins from the pre-trained alexNet by transfer learning. The last three layers of the alexNet are tuned for the score classes. By replacing the last three layers of the alexNet, the network to classify photo scores instead are fine-tuned by feeding the training data set. The generated network is called $aestheticNet_1$. The network architecture is as the following.

   1-22 layers    alexNet layers' Transferring

    23 layer    'fc': full connected layer, N nodes, each corresponding to one class

    24 layer    'softmax': softmax layer, N nodes

25 layer    'classoutput': classification output, '1' and other N-1 classes' crossentropyex

Then, all samples in the training data set are classified into score classes by $aestheticNet_1$, and nodes of 'fc' layer are activated to get the Sigmoidal fuzzy membership values, denoted $fc_{s_n}$. The value of $fc_{s_n}$ of the sample could be treated as its likelihood belong to $s_n$. Based on $fc_{s_n}$, the instances, which apply to be unavailable for the network learning, are dropped out from the original training data set. The idea behind is that the neighboring scores are mutually related, and an instance labeled with the score of the majority classes while having the low likelihood to them may become the imposter nearest neighbors of the minority classes to invade the genuine neighborhood of those. Accordingly, the conditions of dropping-out the unavailable samples are expressed by the expressions (1), (2), and (3).

For a sample $i_{s_{max1}}$, if its $fc_{s_{max1}} < K1$, dropping out  (1)

For a sample $i_{s_{max2}}$, if its $fc_{s_{max1}} < K2$, dropping out  (2)

For a sample $i_{s_{max3}}$, if its $fc_{s_{max1}} < K2$, dropping out  (3)

Where, the score class having most samples is denoted $s_{max1}$, the next is $s_{max2}$, the third is as $s_{max3}$. The sample which is labeled with $s_{max1}$, denoted $i_{s_{max1}}$, is removed from the original training data set if the corresponding $fc_{s_{max1}}$ is less than *K1*. Furthermore, the sample which is labeled with $s_{max2}$ or $s_{max3}$, denoted $i_{s_{max2}}$ or $i_{s_{max3}}$, is dropped out from the original training data set if the corresponding $fc_{s_{max1}}$ is less than *K2*. The values of *K1* and *K2* are adjustable, and the objective to adjusting is to make the number of remaining samples keep about two-thirds of the original data set.

Then, based on the previously trained aesthetic assessment network $aestheticNet_1$, the network is retrained by transfer learning with the new training data set dropping out the unavailable instances. The accordingly generated network is called $aestheticNet_2$. Based on $aestheticNet_2$, the likelihood of all samples of the original training data set belonging to each score class is calculated, and the classification is executed. The samples are dropped out if they apply to the above removing conditions. It is noticed that the samples that are dropped out based on the previously trained network could be remained based on the current network so as to void the sample's miss-removing to loss the valid information.

Repetitively, based on the latest trained aesthetic assessment network $aestheticNet_{i-1}$, the network is retrained by transfer learning with the latest training data set dropping out the unavailable instances. Using the latest generated network $aestheticNet_i$, the likelihood of all samples of the original training data set belonging to each score class is calculated, and classification is executed. The samples are dropped out if they apply to the dropping out conditions. This learning procedure is continued until the value of loss function is less than a threshold *Loss.* Here, the loss function is defined as a quadratic loss function regarding labeling and assigned scores of instances. The finally generated photo aesthetic assessment network is called $aestheticNet_I$.

## 4. Extracting aesthetics highlight region

In this section, we proposed a method of extracting aesthetic highlight regions of photos by using the repetitively trained aesthetic assessment networks, so as to analyze how the photos are assessed by the pro-photographer to investigate the composition features of good photos, and correlations the salient objects with backgrounds in the photos. The diagram of the method is shown in Fig. 3.

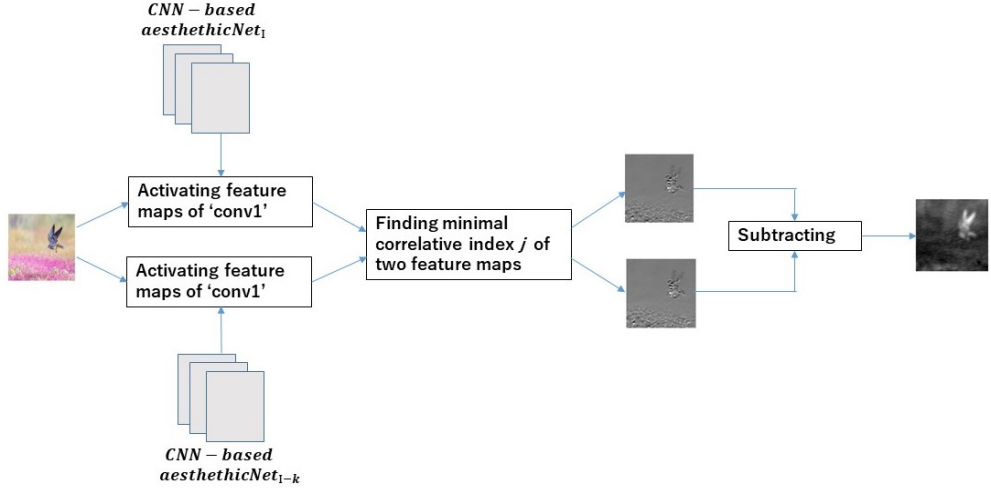

Fig. 3 Diagram of extracting highlight region

For a photo image, feature maps of 'conv1' layer of two CNN-based aesthetic assessment networks are activated. These two networks are the latest retrained network $aestheticNet_I$ and the previously retrained network $aestheticNet_{I-k}$. Two set of the feature maps are denoted by $conv1_I^j$ and $conv1_{I-k}^j$, respectively. *I-k* indicates the *(I-k)th* retrained network, and *j* does the *j*th feature map. It is assumed that the difference map of the two corresponding feature maps, which are of the minimal correlation, could reflect the aesthetics highlight region. The idea behind is that the training data set is aesthetically labeled by the pro-photographer who often focuses on the aesthetics highlight region to embody photo's aesthetic level based on the essential principles, such as whether the object in the photo is distinctive, and whether the composition is concise. So, the activated values of the highlight regions should change more greatly if the network is retrained by removing the unavailable instances. Therefore, the difference map caused by the weights' fine-tuning of networks could be used to extract such aesthetic highlight region.
. The index *J* of the corresponding feature maps which are of the minimal correlation is identified by the following equation.

$$J = \arg min_j (corr[conv_I^j,\ conv_{I-k}^j]) \qquad (4)$$

Where, *corr* indicates the correlation. The difference map of the two feature maps with index *J* could be calculated by subtracting $conv1_I^J$ and $conv1_{I-k}^J$, the equation of which is as the following.

$$diff_{I,I-k} = \text{conv}1_I^J - \text{conv}1_{I-k}^J \tag{5}$$

Fig.4 shows the difference maps of two photos. The left are original images; the middle are feature maps regarding $\text{conv}1_{10}^J$ and $\text{conv}1_9^J$, respectively; the right are the difference maps of them, denoted $diff_{10,9}$.

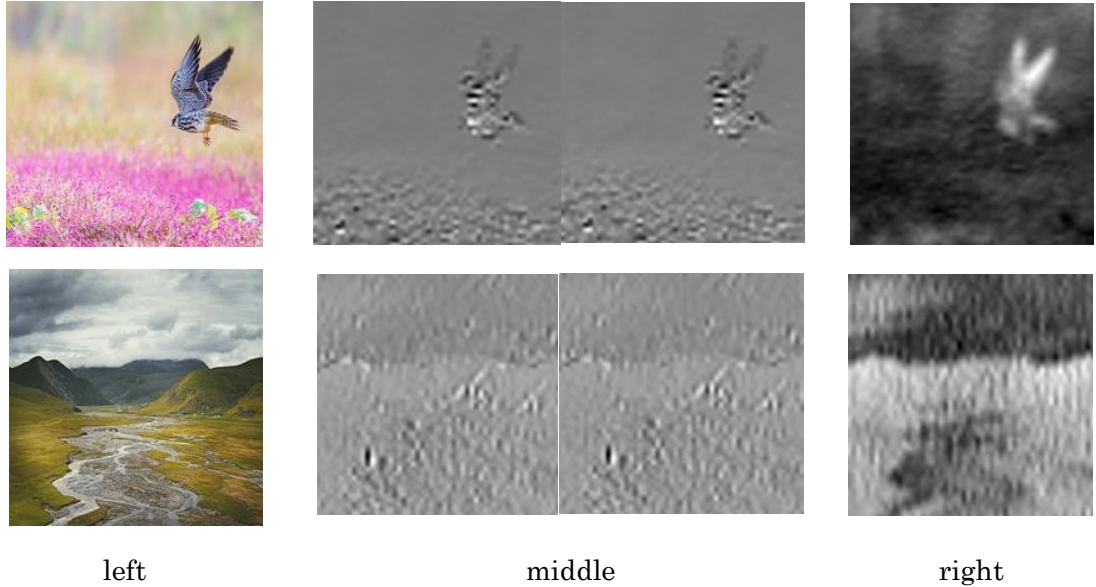

left            middle            right

Fig. 4 Calculating different map

For the upper instance, the salient object bird as a highlight is emerged explicitly in the difference map; for the lower instance, the mountain area is emerged in the different map although the highlight of this insatnace is not obvious.

Moreover, the different maps of above two instances regarding $\text{conv}1_2^J$ and $\text{conv}1_1^J$ is shown in Fig.5. $\text{conv}1_1^J$ is the feature map of $aestheticNet_1$ which is firstly trained based on alexNet by the original data set.

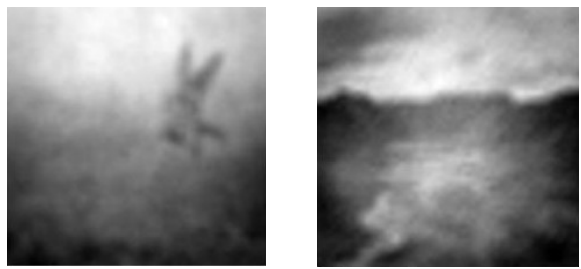

Fig.5 different maps regarding $\text{conv}1_2^J$ and $\text{conv}1_1^J$

The left is the result of the upper instance, and the right corresponds to the lower instance. Obviously, the highlight regions aren't emerged explicitly in these two difference maps. So, it is indirectly validated the effectiveness of repetitively training network by dropping out the unavailable instances.

## 5. Experiments and analysis

### 5.1 Data set

Although the AVA data set [12] is the largest publicly available aesthetics dataset providing over 250,000 images in total, each image in which was aesthetically assessed by about 200 people with the rating score ranging from 1 to 10, all of the images were finally labeled with the mean score that lost the individual`s aesthetic sense although the aesthetic tendencies of a human group could be reflected. However, embodying the aesthetic taste of a human being is important in training our aesthetic assessment model. Therefore, we conduct our training data set which contains 3100 photos assessed aesthetically by a professional photographer, which is called xiheAA. These photos were taken by the students of the photographer's class. The scores ranging from 2 to 9. The distribution of the scores is shown in Fig. 1. The class having most samples is score 4; the next is score 3; the third is score 5. Also, we collect testing data from the other source, 500px. 310 images were downloaded randomly from the 500px's recommended photo set.

### 5.2 Aesthetic level classification

In this section, we elaborate how the repetitively trained network models improve the aesthetic level classification performance. The method proposed in section 3 is implemented on Matlab. The function *trainNetwork* is used to fine-tune the hyper-parameters of the network based on transfer learning. The function *classify* is used to classify the samples based on the trained network.

Table 1 illustrates the different score class's classification accuracies of the samples in the original xiheAA data set against the different networks. As described above, $aestheticNet_1$ is the network fine-tuning the hyper-parameters of alexNet by transfer learning using the original xiheAA data set. $aestheticNet_i$ *(i>=2)* is the network fine-tuning the hyper-parameters of $aestheticNet_{i-1}$ by transfer learning using the new xiheAA data set from which the samples in the classes of score 3, score 4 and score 5 that fulfill the conditions expressed by (1), (2) and (3) are dropped out. *K1* and *K2* are ranged from 0.85 to 0.95 to make the removed samples keep about one-thirds of the original data set.

Table 1 Score's Classification accuracies against different networks

| Score | $aestheticNet_1$ | $aestheticNet_2$ | $aestheticNet_3$ | $aestheticNet_9$ | $aestheticNet_{10}$ |
|---|---|---|---|---|---|
| 2 | 0.58 | 0.88 | 0.96 | 1 | 1 |
| 3 | 0.46 | 0.54 | 0.75 | 0.9 | 0.92 |
| 4 | 0.94 | 0.78 | 0.82 | 0.91 | 0.91 |
| 5 | 0.5 | 0.73 | 0.73 | 0.84 | 0.85 |
| 6 | 0.73 | 0.82 | 0.87 | 0.99 | 0.99 |
| 7 | 0.33 | 0.59 | 0.78 | 0.86 | 0.86 |
| 8 | 0.21 | 0.64 | 1 | 1 | 1 |
| 9 | 0 | 0.5 | 1 | 1 | 1 |

It is observed from the results shown in Table 1 that the classification accuracies of minority classes with salient features, such as the class of score 2 indicating poor aesthetic quality, and the classes of score 7, score 8, score 9 indicating excellent quality, are improved dramatically by using the repetitively trained networks. In the cases of score 2, score 8 and score 9, the accuracies increase gradually from low levels to 100%. In the case of score 7 with the relatively high number of samples, the accuracies are improved from 33% to 86%. For the relatively majority classes of score3 and score 5, the classification accuracies also ameliorate gradually from low levels to high levels. For the class of score 4 with the dominant samples, the classification accuracy is 94% by using the initial network $aestheticNet_1$. Then the accuracy decreases to 78% using the first retrained network $aestheticNet_2$. And then, the accuracies increases gradually using the further retrained networks. The accuracy becomes 91% using the retrained network $aestheticNet_{10}$, although this doesn't reach to the initial accuracy 94%. The changing tendency of the accuracies is reasonable because removing the unavailable instances in the domination classes in training ameliorates the invasion of these samples to the minority classes so as to achieve the balance of the classification accuracies against the different classes. Table 2 shows the Loss values against the repetitively trained different networks.

Table 2 Loss values against different networks

|  | $aestheticNet_1$ | $aestheticNet_2$ | $aestheticNet_3$ | $aestheticNet_9$ | $aestheticNet_{10}$ |
|---|---|---|---|---|---|
| Loss | 2.27 | 2.2 | 2.15 | 1.8 | 1.77 |

It is observed that the loss value decrease gradually with repetitively training the network. The change tendency corresponds to the increasing of the classification accuracy shown in Table 1.

Next, the testing data set that were downloaded randomly from the 500px's recommended photo set is used to verity the effectiveness of the proposed method.

Because these samples were selected by pro-photographers, it is thought that the qualities of aesthetic level are relatively high. Many of them should be over the fair levels, although they are not scored. So, the assigned rate is conducted to validate the model's aesthetic level classification performance against such testing data set. The assigned rate is defined by equation (6).

$$assignRate_i = \frac{\#assign_i}{\#all} \quad (6)$$

Where, # indicates the number, $\#assign_i$ indicates the number of photo images assigned to class *i*, and $\#all$ indicates the number of photos in the data set. Table 3 shows the assigned rates of the samples in the testing data set using the different networks. $aestheticNet_1$ is the initial aesthetic assessment network, and $aestheticNet_i$ (i>=2) is the retrained network by the data set dropping out the unavailable instances. Table 3 illustrates the assigned rates of the samples to the different score classes.

Table 3 assigned rates of the samples

| Score | $aestheticNet_1$ | $aestheticNet_2$ | $aestheticNet_3$ | $aestheticNet_9$ | $aestheticNet_{10}$ |
|---|---|---|---|---|---|
| 2 | 0 | 0 | 0 | 0 | 0 |
| 3 | 0.29 | 0.16 | 0.16 | 0.17 | 0.15 |
| 4 | 0.46 | 0.34 | 0.38 | 0.32 | 0.28 |
| 5 | 0.14 | 0.3 | 0.21 | 0.22 | 0.26 |
| 6 | 0.046 | 0.07 | 0.1 | 0.1 | 0.11 |
| 7 | 0.054 | 0.12 | 0.15 | 0.19 | 0.2 |
| 8 | 0 | 0 | 0 | 0 | 0 |
| 9 | 0 | 0 | 0 | 0 | 0 |

It is observed that the assigned rates of the samples to the score classes is dramatically changed by using the retrained networks, especially to the classes with salient features. For example, for the class of score 7 indicating the excellent aesthetic level, the assigned rate by using the network $aestheticNet_{10}$ is about four times to one by using the initial $aestheticNet_1$. For the classes of fair level with the score less than 5, the assigned rates decrease about 50%. For the classes of score 5 and score 6 indicating good level, the rates increase about 200%. The total assigned rate of the classes with the score over 4 increases 238% from 0.24 to 0.57. Such result is acceptable, because it is accordant to the above assumption that the aesthetic quality of most of the samples in the testing data set should be good.

On the other hand, some samples assigned to the classes of score 3 and score 7 are shown in Fig. 6. The upper images are assigned to the score 3, and the lower images are assigned to score 7.

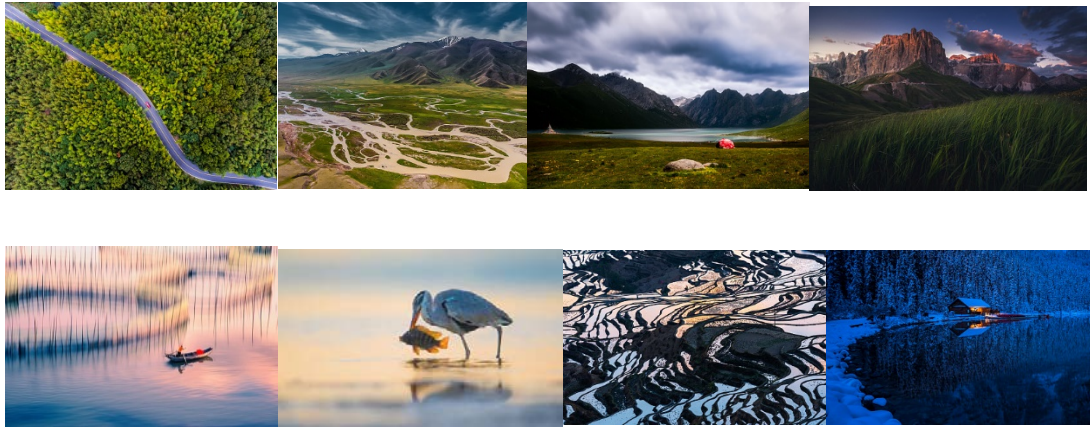

Fig.6 some samples assigned to the classes of score 3 and score 7

It is obviously that the visual aesthetic quality of the lower images is better than one of the upper images, and seems to meet the common techniques for composing a good photo. So, the classification results of these samples are reliable. Moreover, it make us know that the fair photos exist in the above testing data set although they are selected by the pro-photographers. Accordingly, it is verified that the results shown in Table 3 are acceptable.

As a whole, the experimental results demonstrates that the proposed method for the photo's aesthetic level classification is effective.

### 5.3 Aesthetic analysis of highlight region

In section 4, extracting aesthetics highlight region from the photo image by using the repetitively trained networks was proposed. In this section, we focus on analyzing the correlation of the extracted region with the aesthetic assessment, so as to illustrate how to improve the photo's aesthetic quality.

Fig.7 shows the aesthetics highlight regions of some photos labeled with score 2 in the training data set. The left is the set of some original images; the middle is the set of the corresponding different maps calculated by $aestheticNet_{10}$ and $aestheticNet_9$; the right is the set of the extracted highlight regions based on the different maps using the method of 2-means.

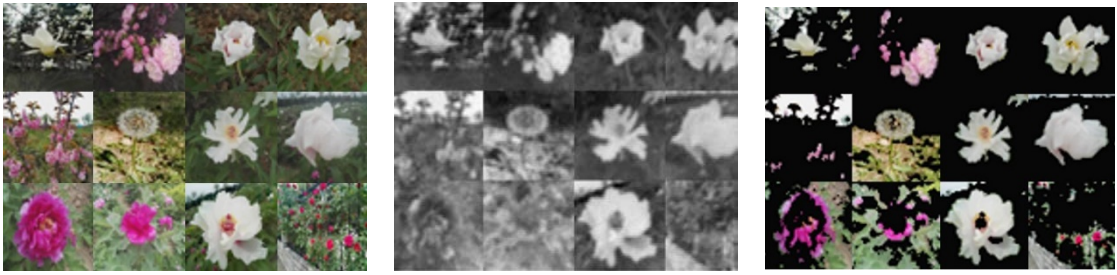

Fig. 7 highlight regions of some photos labeled with score 2

It is obviously that the extracted parts are messed and cluttered. The salient objects look ugly. Of course, the photo's highlight region with such features make the aesthetic assessment bad.

Fig. 8 shows the highlight regions of some photos labeled with score 4 in the training data set.

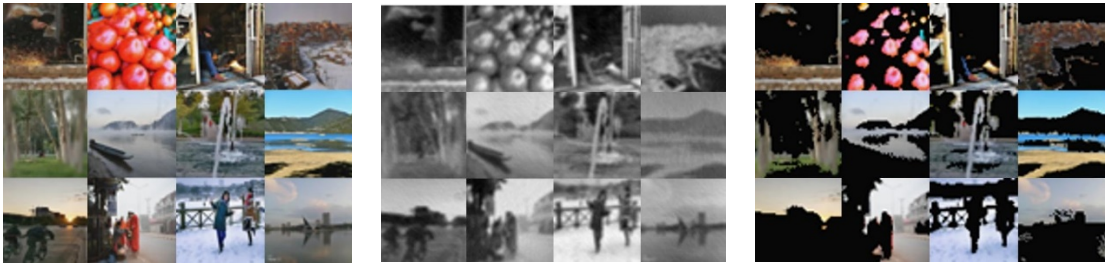

Fig. 7 highlight regions of some photos labeled with score 4

It is observed that the extracted parts look plain and dull. There are not the salient objects in the extracted region. So, we know that the photos without salient objects often obtain the fair assessment.

Fig. 9 shows the highlight regions of some photos labeled with score 7 in the training data set.

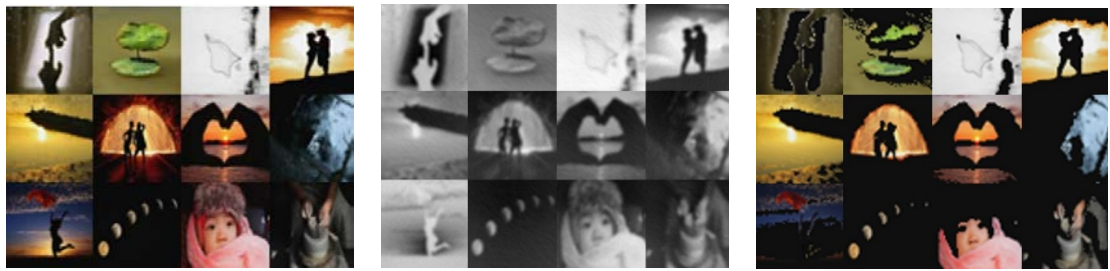

Fig. 9 shows the highlight regions of some photos labeled with score 7

It is observed that the extracted parts are very clear. The salient objects are distinctive and made outstanding, and look pretty. So, we understand that the distinctive object with the clear highlight region make the photo have the high aesthetic quality.

Accordingly, we can say that the photos' aesthetics highlight regions extracted by using the repetitively trained aesthetic assessment network reveal the photo's aesthetic qualities. By analyzing the compositions of the extracted elements with the aesthetic scores assigned to the photos, it is possible to learn how to arrange the elements in the photo to make up a good photo.

## 6. Conclusion

In this paper, we propose a method of learning the aesthetic assessment model with the imbalance data set. Using CNN and transfer learning, the aesthetic network was repetitively trained by dropping out the unavailable instances in the training data set. The mechanism of relearning and the conditions of removing was presented, and how the repetitively trained network models improve the aesthetic level classification performance was analyzed. Moreover, based on the feature maps of the retrained models, a method of extracting aesthetics highlight region of the photo was described, and how such region correlates to the aesthetic assessment was illustrated. Overall, the system could not only estimates the photo's aesthetic level, but also indicate why this estimation was induced. We think that the proposed method is also available for other domains which are relevant to the subjective assessment.

## Reference

[1] ColinG. Johnson Jon McCormack, Iria Santos, and Juan Romero, "Understanding Aesthetics and Fitness Measures in Evolutionary Art Systems", Hidawi Complexity, Volume 2019, Article ID 3495962, https://doi.org/10.1155/2019/3495962

[2] Yubin, Chen Change Loy, and Xiaoou Tang, "Image aesthetic assessment: an experimental survey", arXiv: 1610.00838v2

[3] Eftichia Mavridaki, Vasileios Mezaris, "A comprehensive aesthetic quality assessment method for natural images using basic rules of photography", 2015 IEEE International Conference on Image Processing (ICIP) , Canada, 2015

[4] Zhe Dong, Xinmei Tian, "Multi-level photo quality assessment with multi-view features", Neurocomputing, Volume 168, 30 November 2015, Pages 308-319

[5] Yueying Kao, et al. "Hierarchical aesthetic quality assessment using deep convolutional neural networks", Signal Processing: Image Communication, Volume 47, September 2016, Pages 500-510

[6] Weining Wang, et al. "A multi-scene deep learning model for image aesthetic evaluation", Signal Processing: Image Communication, Volume 47, September 2016, Pages 511-518,

[7] Yunlan Tan, et al. "Photograph aesthetical evaluation and classification with deep convolutional neural networks", Neurocomputing, Volume 228, 8 March 2017, Pages 165-175

[8] François Lemarchand, "Fundamental visual features for aesthetic classification of photographs across datasets", Pattern Recognition Letters, Volume 112, 1 September 2018, Pages 9-17

[9] Chen Huang, et al. " Learning Deep Representation for Imbalanced Classification", 2016 IEEE Conference on Computer Vision and Pattern Recognition (CVPR), DOI: 10.1109/CVPR.2016.580, USA, 2016

[10] Chao Zhang, et al. "Visual aesthetic understanding: Sample-specific aesthetic classification and deep activation map visualization", Signal Processing: Image Communication, Volume 67, September 2018, Pages 12-21

[11] H. He and E.A. Garcia, "Learning from imbalanced data", TKDE, 21(9), Pages 1263-1284, 2009

[12] N. Murray, et al. "AVA: a large-scale database for aesthetic visual analysis", Proceedings of the IEEE conference on computer vision and pattern recognition, 2012, Pages 2408-2415